\documentclass[a4paper, oneside, twocolumn, notitlepage, 10pt]{extarticle_ecoc}
\usepackage{ecoc}

\usepackage{subcaption}
\usepackage{caption}
\usepackage{comment}

\begin{document}
\selectlanguage{english}    


\title{
Experimental Demonstration of Event-based Optical Camera Communication in Long-Range Outdoor Environment
}%


\author{
    Miu Sumino\textsuperscript{(1)},
    Mayu Ishii\textsuperscript{(1,2)},
    Shun Kaizu\textsuperscript{(2)},
    Daisuke Hisano\textsuperscript{(3)},
    Yu Nakayama\textsuperscript{(1)}
}

\maketitle                  


\begin{strip}
    \begin{author_descr}

        \textsuperscript{(1)}Department of Computer and Information Sciences,
        Tokyo University of Agriculture and Technology, Tokyo, Japan
        \textcolor{blue}{\uline{\{miu.sumino, yu.nakayama\}@ynlb.org}}

        \textsuperscript{(2)}Sony Semiconductor Solutions Corporation, Tokyo, Japan
        
        \textsuperscript{(3)}Graduate School of Engineering, The University of Osaka, Osaka, Japan

    \end{author_descr}
\end{strip}

\renewcommand\footnotemark{}
\renewcommand\footnoterule{}


\begin{strip}
    \begin{ecoc_abstract}
        We propose a robust demodulation scheme for optical camera communication systems using an event-based vision sensor, combining OOK with toggle demodulation and a digital phase-locked loop. This is the first report to achieve a BER < {10\textsuperscript{-3}} at 200m-60kbps and 400m-30kbps in outdoor experiments.\textcopyright2025 The Author(s)
    \end{ecoc_abstract}
\end{strip}

\begin{figure*}[!b]
  \centering
  \begin{subfigure}[t]{0.7\linewidth}
    \includegraphics[width=\linewidth]{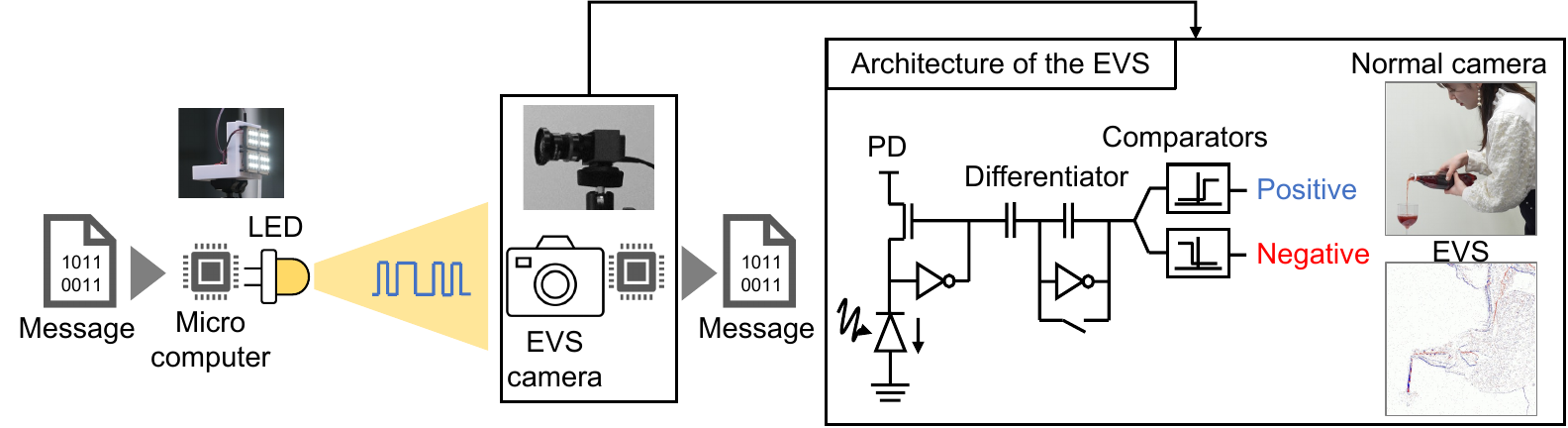}
    \caption{Overview.}\label{fig:evs}
  \end{subfigure}
  \hfill
  \begin{subfigure}[t]{0.29\linewidth}
    \includegraphics[width=0.8\linewidth]{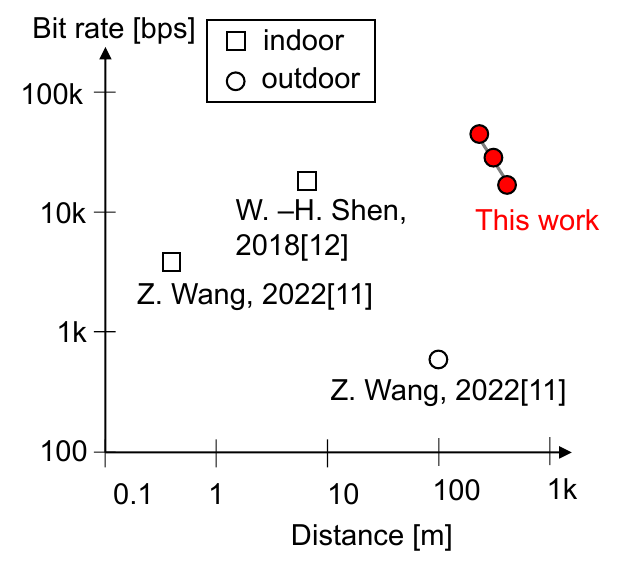}
    \caption{Benchmark of OCC Systems.}\label{fig:bench}
  \end{subfigure}
  \vspace{4mm}
  \caption{EVS-based OCC System.}
\end{figure*}

\section{Introduction}
Optical camera communication (OCC) is a form of visible light communication (VLC), in which a complementary metal-oxide-semiconductor (CMOS) sensor-based camera is used as the receiver to capture modulated light emitted from light sources such as LED lights, panels, displays, and digital signage\cite{le2017survey,seminara2021characterization,chen2017oled,higuchi2024cipher}. One of the key advantages of OCC is that it operates in the visible light spectrum, which does not overlap with radio frequency (RF) bands such as Wi-Fi or Bluetooth\cite{hasan2021optical,vappangi2019concurrent}. This allows OCC to function without causing electromagnetic interference, making it suitable for environments where RF usage is restricted, such as in aircraft cabins and medical facilities\cite{ali2022recent}.

In conventional OCC systems, frame-based cameras are used as receivers\cite{luo2018undersampled, rachim2018multilevel}. This introduces several limitations. For example, the achievable throughput is limited by the frame rate of the camera\cite{chia2023high}, and the large volume of image data captured requires significant computational resources for real-time processing.

To address these challenges, OCC systems that use event-based vision sensors (EVS) as receivers have been proposed\cite{wang2022smart, shen2018vehicular, nakagawa2024linking} as shown in Fig. \ref{fig:evs}. Unlike traditional frame-based cameras, EVS operates asynchronously and detects changes in brightness at the pixel level\cite{gallego2020event,scheerlinck2018continuous,lichtsteiner2008128}. As illustrated in Fig. \ref{fig:evs}, the EVS outputs only the information related to pixels where brightness changes occur, together with their spatial coordinates and timestamps. Note that in the following sections, OCC systems that employ EVS as the receiver are referred to as event-based OCC.

In previous works on event-based OCC, modulation schemes such as on-off keying (OOK) \cite{wang2022smart} and pulse position modulation (PPM) \cite{shen2018vehicular} have been applied to modulate the optical signals. A summary of the transmission capacity and distance reported in these works is shown in Fig. \ref{fig:bench}.
Although there are reports achieving data rates over 100 kbps\cite{su2024motion}, they are excluded from this discussion because they employ multi-channel configurations and exhibit bit error rate (BER) as high as {10\textsuperscript{-2}}.

This paper aims to further enhance the transmission capacity of event-based OCC. To this end, we optimize the frequency response of the EVS to match the frequency of the transmitted optical pulses. Additionally, a digital phase-locked loop (DPLL) is introduced to compensate for burst errors caused by missed pulse detection. Notably, our system employs a low-cost microcontroller, such as an Arduino and M5Stack, to drive the LED-based transmitter. Although temporal jitter unavoidably arises due to the use of such devices, the receiver-side DPLL effectively mitigates its impact, thereby suppressing the degradation of BER.

Finally, we report the results of outdoor transmission experiments, in which our proposed system achieved record-setting performance. Figure \ref{fig:bench} presents a comparison of transmission distance and capacity between previous studies and our method. As the proposed system successfully achieved both longer transmission distance and higher bit rate compared to conventional approaches.


\section{EVS Characteristics}
The pixel architecture of the EVS is illustrated as follows. Incident light is first converted into an analog voltage by a photodiode. This analog signal is then amplified and passed through tunable filters, such as lowpass or high-pass filters that shape the temporal response of the pixel. The filtered signal is fed into a differentiator circuit, which detects changes in luminance. When the amount of change exceeds a predefined contrast threshold, a spiking event is generated and output. Several parameters can be individually tuned for each pixel, including the filter bandwidth, refractory period, and contrast sensitivity. By optimizing these parameters, the temporal resolution and responsiveness of the sensor can be significantly improved.

When applied to an OCC system, the high-frequency switching of the LED causes transient analog responses that often result in multiple events per pixel, with slight variations across different pixels. To address this, the events from relevant pixels are aggregated and employed a smoothing filter on. This process yields a continuous waveform that accurately reflects the blinking pattern of the LED.
\section{Principle of Proposed Event-based OCC}
\begin{figure}
    \centering
    \includegraphics[width=\linewidth]{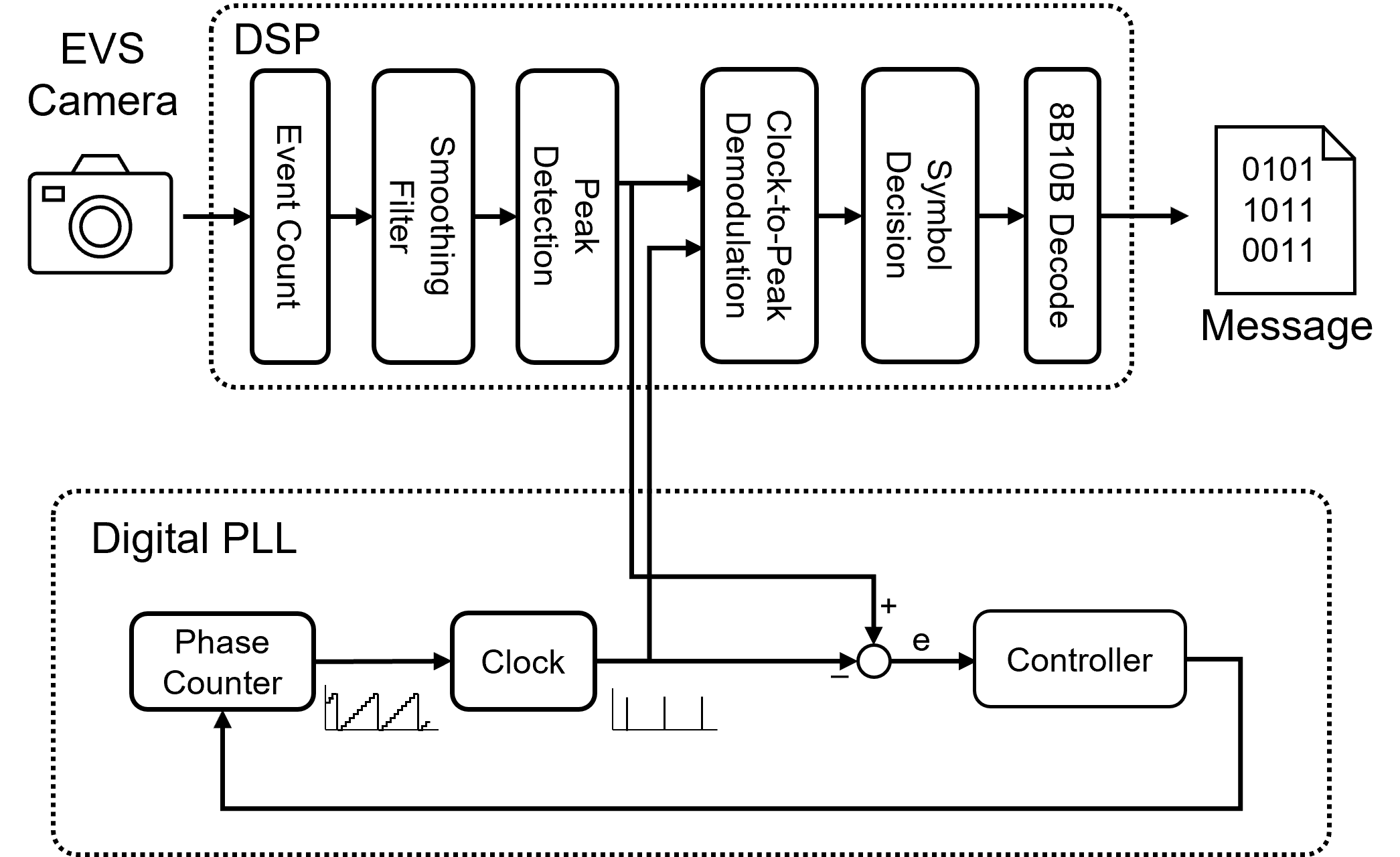}
    \caption{Receiver Configuration.}
    \label{fig:flow}
\end{figure}

\begin{figure}
    \centering
    \includegraphics[width=\linewidth]{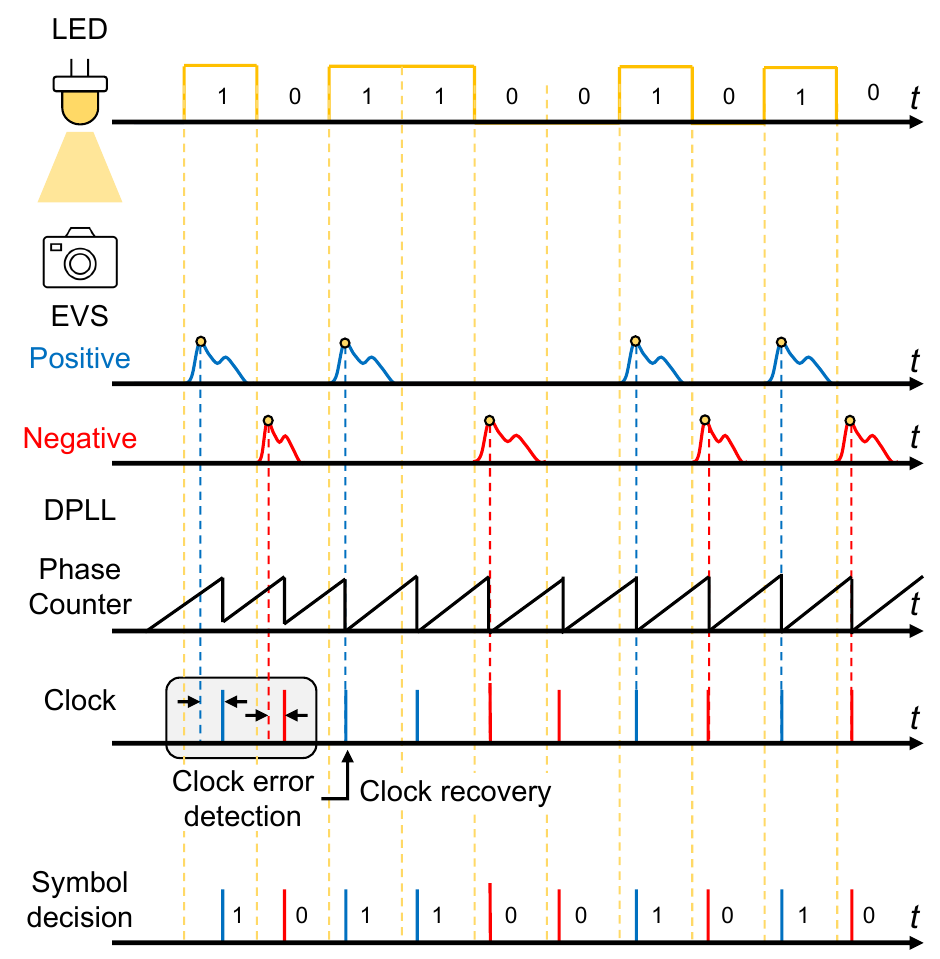}
    \caption{Operation Sequence.}
    \label{fig:time}
\end{figure}

This paper proposes a modulation and demodulation scheme for an OCC system using an EVS, which combines OOK modulation with toggle-based demodulation. The functional blocks and operation sequence of the proposed event-based OCC system are illustrated in Fig. \ref{fig:flow} and Fig. \ref{fig:time}.

On the transmitter side, a continuous non-return-to-zero (NRZ) OOK signal is transmitted. For reliable clock extraction, it is desirable to employ a line coding scheme such as 8b/10b. On the receiver side, the EVS detects changes in light intensity by outputting positive and negative events. Since the detected events may be distorted due to external noise and disturbances, a smoothing filter is applied to facilitate peak detection. Subsequently, the peaks of the events are extracted.
When a peak in the positive events is detected and no negative event is observed for a certain duration, the bit state is considered to remain at “1.” Conversely, when a peak in the negative events is detected, the bit state transitions from “1” to “0.” This behavior can be interpreted as a state-holding mechanism similar to that of a flip-flop (FF), where positive events function as set pulses and negative events as reset pulses. In this paper, we refer to this demodulation scheme as toggle demodulation.
As positive and negative events are generated asynchronously, timing jitter occur in the intervals between these events. Such jitter can result in missing bits, which in turn causes burst errors. To suppress the impact of this jitter, we introduce a digital phase-locked loop (DPLL) in addition to toggle demodulation.

\section{Experiments}

\begin{figure}[t]
    \centering
    \includegraphics[width=0.7\linewidth]{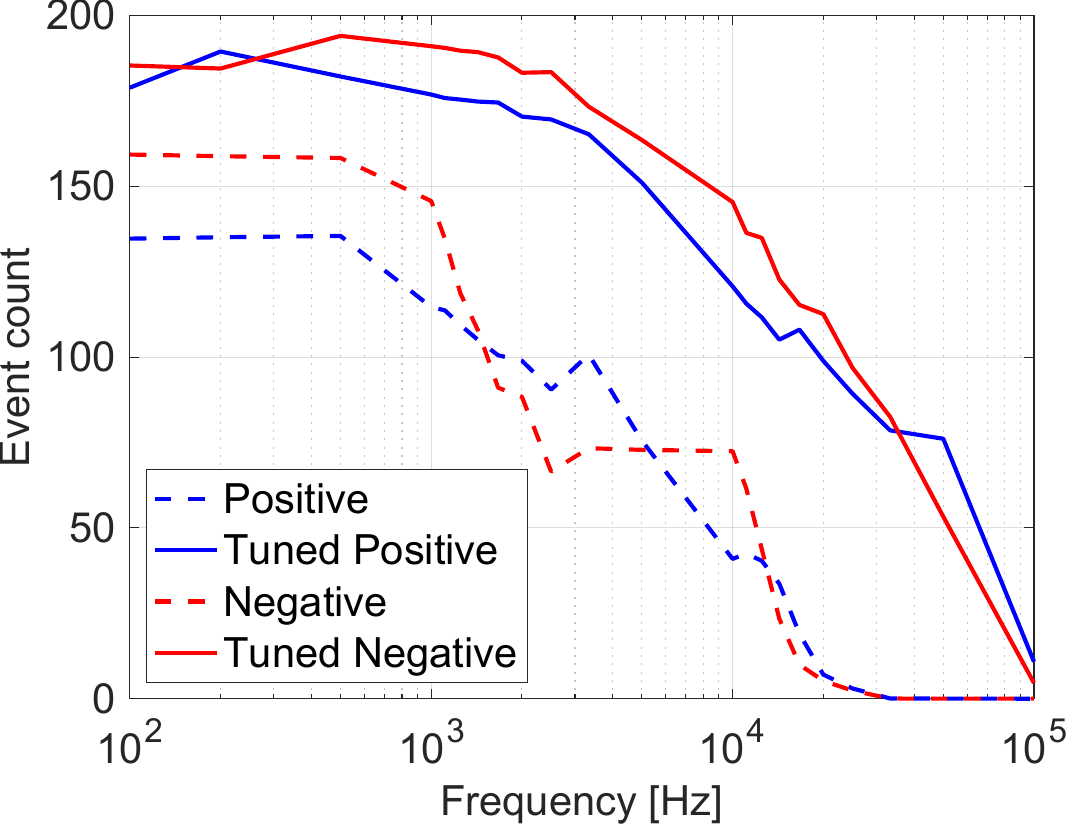}
    \caption{Frequency Response.}
    \label{fig:freq}
\end{figure}

\begin{figure}[t]
    \centering
    \includegraphics[width=0.8\linewidth]{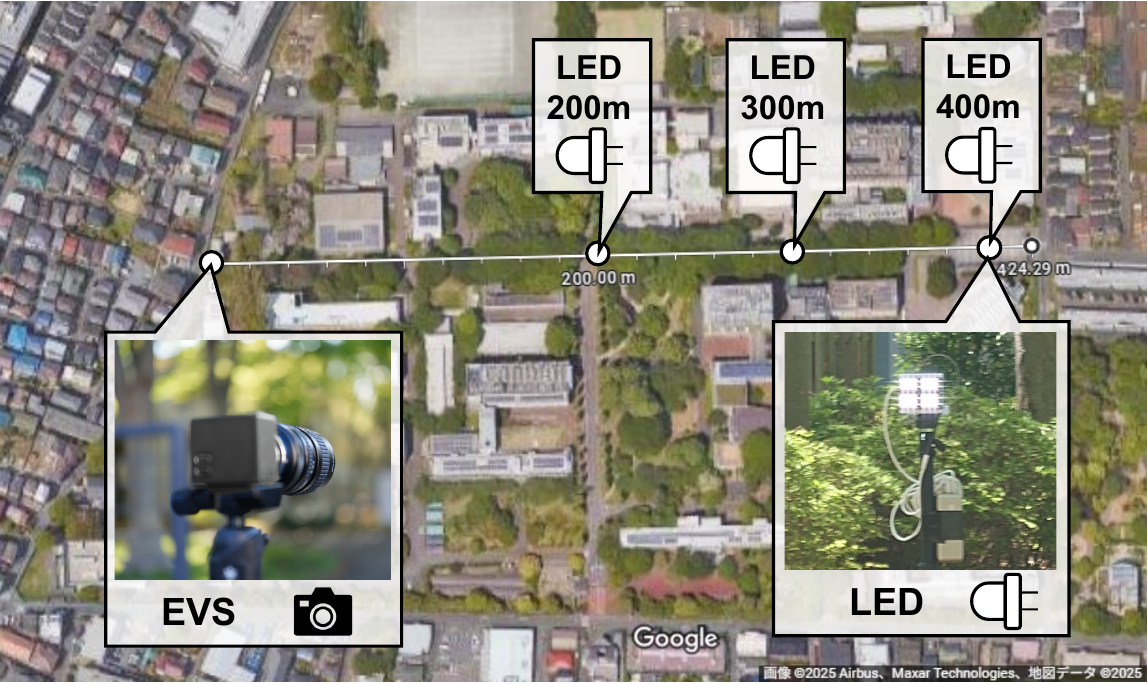}
    \caption{Experimental Setup.}
    \label{fig:setup}
\end{figure}

The experimental setup is shown in Fig.~\ref{fig:setup}.
The M5Stack modulated these symbols and controlled the blinking of the LED. This experiment utilized surface mounting chip LEDs manufactured as the transmitter, which is white in color and has a brightness of 26.5 lm. 
We employed a 0.92-megapixel EVS, the IMX636 from Sony Semiconductor Solutions, as the receiver, and equipped it with a RICOH FA lens featuring a 75 mm focal length and an F1.8 aperture.
To reduce computational load, only the pixels capturing the area around the light source were cropped and used, instead of processing the entire pixel array.
We calibrated each parameter of the EVS and evaluated its frequency characteristics by varying the on/off frequency of the LED. Figure~\ref{fig:freq} illustrates the number of events observed at each frequency. Parameter tuning improved event occurrence at high frequencies above 10 kHz for both positive and negative signals, compared to the initial settings.

In the demodulator, the number of events was counted, a smoothing filter was applied, and the intervals between local maximums were calculated as shown in Fig.~\ref{fig:flow}. Subsequently, demodulation was performed, and the original bit sequence was recovered. All of these processes on the receiver side were conducted through offline signal processing.

Figure~\ref{fig:setup} shows the experimental location. The experiment was conducted on the Koganei Campus of the Tokyo University of Agriculture and Technology, located in Tokyo, Japan. It was performed outdoors during the daytime at transmission distances of 200 m, 300 m, and 400 m. The environment included shadows and intermittent cloud cover, with an illuminance of at least 10 klux.

A bit stream consisting of 8 kbit random codes encoded in 8b/10b is prepared and blinking with LED.
By transmitting a total of 10 kbit codes at least 100 times, the BER is calculated.
Transmissions were conducted at each of the three distances while varying the on/off frequency from 10 kHz to 100 kHz.

\begin{figure}[!t]
    \centering
    \includegraphics[width=0.86\linewidth]{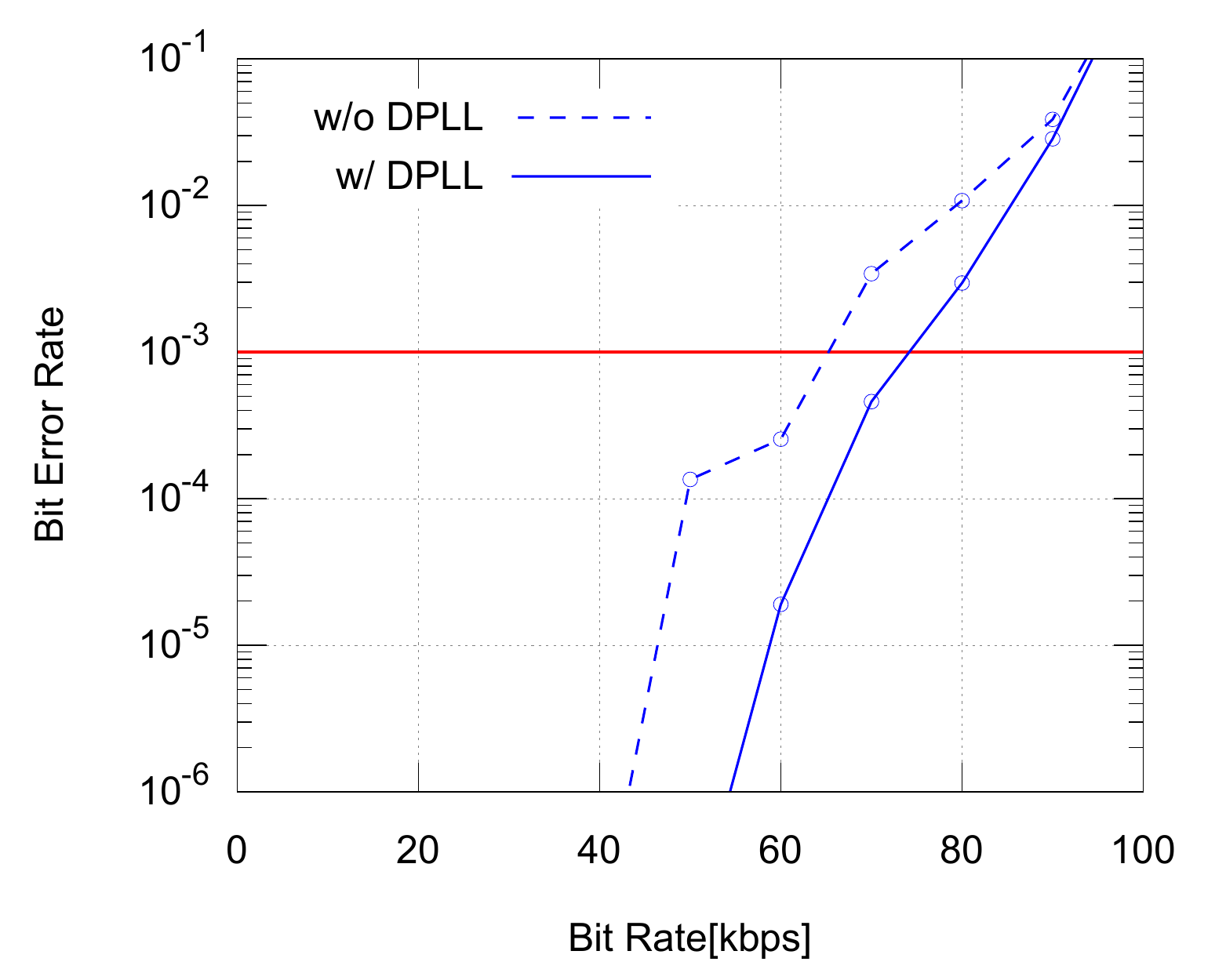}
    \caption{Bit Error Rate of 200m.}
    \label{fig:result200}
\end{figure}
\begin{figure}[!t]
    \centering
    \includegraphics[width=0.86\linewidth]{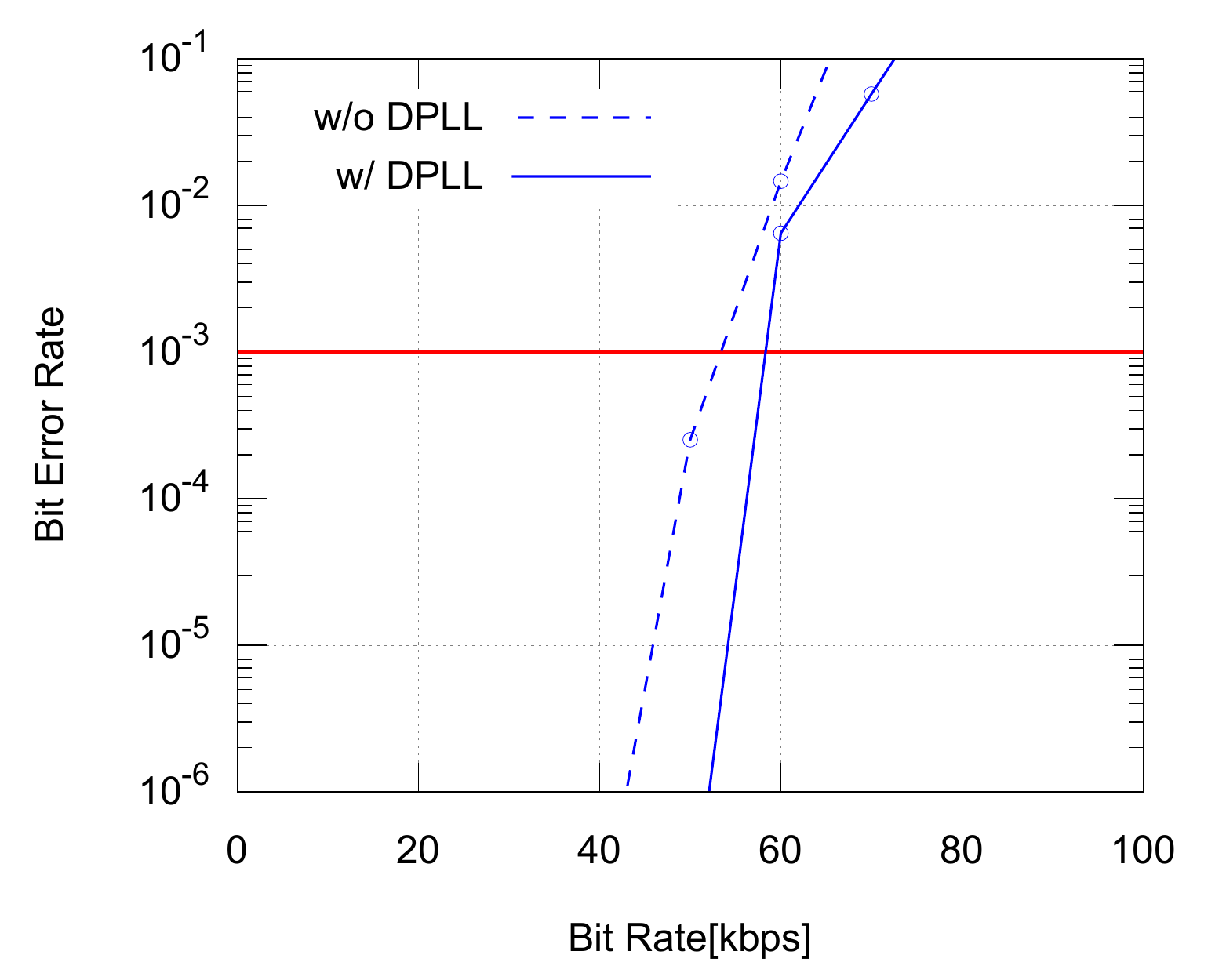}
    \caption{Bit Error Rate of 300m.}
    \label{fig:result300}
\end{figure}
\begin{figure}[!t]
    \centering
    \includegraphics[width=0.86\linewidth]{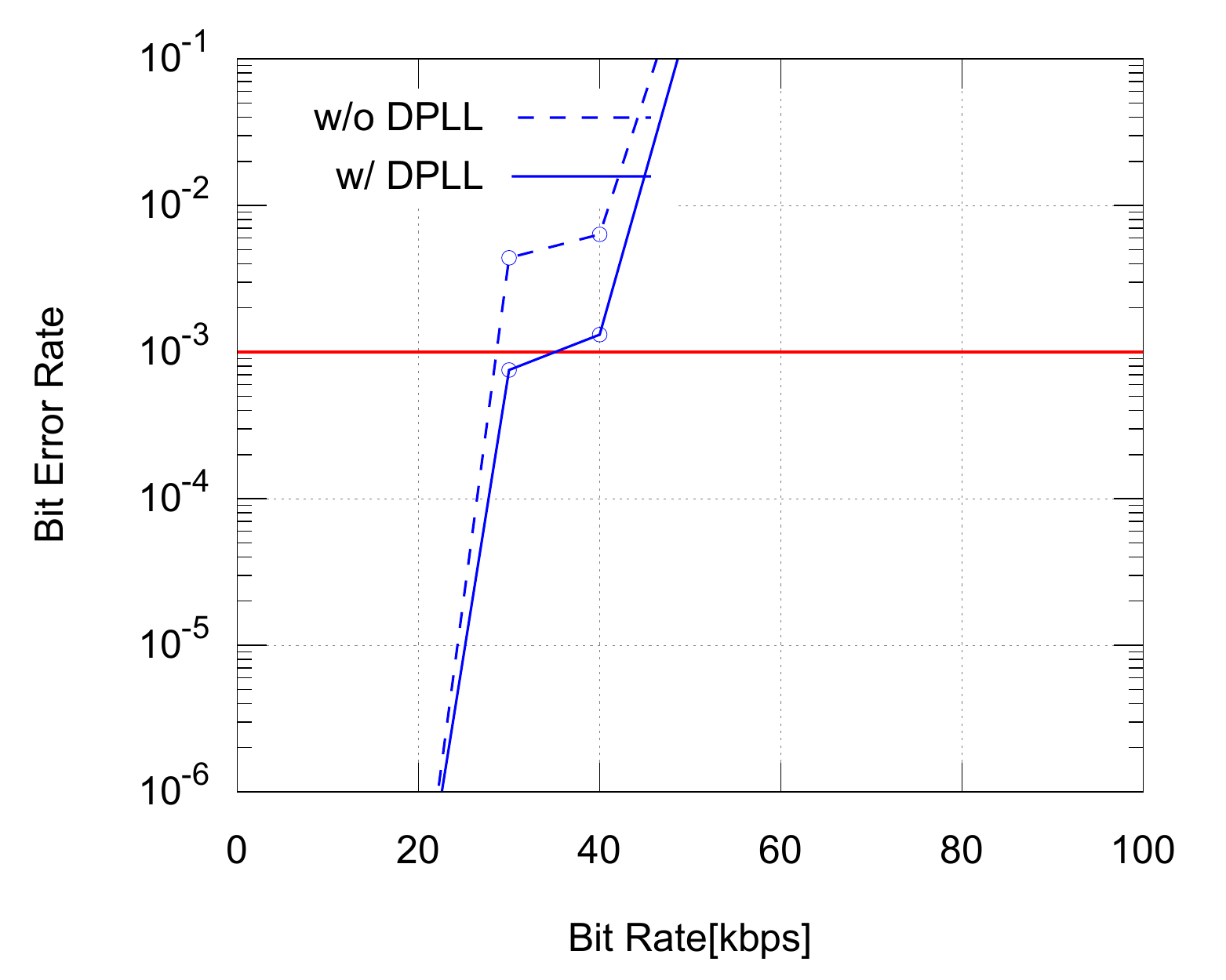}
    \caption{Bit Error Rate of 400m.}
    \label{fig:result400}
\end{figure}

Figure~\ref{fig:result200}-\ref{fig:result400} shows the experimental results. We achieved transmission capacity of 60 kbps, 50 kbps, and 20 kbps at the respective distances by allowing a less than BER of {10\textsuperscript{-3}}, which is within 7\% hard-decision forward error correction (HD-FEC) limit.
To verify the effectiveness of the DPLL, we compared scenarios with and without the DPLL. At each transmission distance, the BER extremely decreased when it was applied, indicating that the use of DPLL improved demodulation stability.




\section{Conclusion}
This paper proposed a new demodulation technique combining OOK modulation with toggle demodulation and a DPLL for event-based OCC systems. The proposed method demonstrated that even when using inexpensive microcontrollers such as Arduino or M5Stack for the transmitter, it was possible to maintain stable long-distance communication and achieve high data rates. Outdoor long-range transmission experiments were conducted, achieving data rates of 60 kbps at 200 m, 50 kbps at 300 m, and 20 kbps at 400 m. After accounting for the overhead of 8b/10b encoding and a 7\% loss due to HD-FEC, the net data rates were 44.6 kbps, 37.2 kbps, and 14.9 kbps at each distance, significantly outperforming conventional methods. Future work will focus on real-time implementation.

\clearpage
\printbibliography

\vspace{-4mm}

\end{document}